\documentclass{article}

\usepackage[preprint]{neurips_2025}

\usepackage[utf8]{inputenc} 
\usepackage[T1]{fontenc}    
\usepackage{hyperref}       
\usepackage{url}            
\usepackage{booktabs}       
\usepackage{amsfonts}       
\usepackage{nicefrac}       
\usepackage{microtype}      
\usepackage{xcolor}         
\usepackage{graphicx}

\title{Accelerate Creation of Product Claims Using Generative AI}

\author{%
  Po-Yu Liang\\
  Department of Computer Science\\
  University of Cincinnati\\
  Cincinnati, OH 45219 \\
  \texttt{liangpu@mail.uc.edu} \\
  \And
  Yong Zhang \\
  Corporate Functions R\&D, Discovery\& Innovation Platforms, P\&G \\
  Mason, OH 45040 \\
  \texttt{zhang.y.13@pg.com} \\
  \AND
  Tatiana Hwa \\
  Beauty Care IT, P\&G \\
  Mason, OH 45040 \\
  \texttt{hwa.t@pg.com} \\
  \And
  Aaron Byers \\
  Beauty Care R\&D, P\&G  \\
  Mason, OH 45040 \\
  \texttt{byers.al@pg.com} \\
}

\begin{document}

\maketitle

\begin{abstract}
The benefit claims of a product is a critical driver of consumers' purchase behavior. Creating product claims is an intense task that requires substantial time and funding. We have developed the \textbf{Claim Advisor} web application to accelerate claim creations using in-context learning and fine-tuning of large language models (LLM). \textbf{Claim Advisor} was designed to disrupt the speed and economics of claim search, generation, optimization, and simulation. It has three functions: (1) semantically searching and identifying existing claims and/or visuals that resonate with the voice of consumers; (2) generating and/or optimizing claims based on a product description and a consumer profile; and (3) ranking generated and/or manually created claims using simulations via synthetic consumers. Applications in a consumer packaged goods (CPG) company have shown very promising results. We believe that this capability is broadly useful and applicable across product categories and industries. We share our learning to encourage the research and application of generative AI in different industries.

\end{abstract}

\section{Introduction}

Product claims play a crucial role in influencing consumer choice and building brand trust, as they provide essential information that helps both consumers and healthcare professionals assess the suitability, efficacy, and safety of various products \cite{couste2012power}. Clear, well-substantiated claims can significantly enhance consumer confidence, serving as a critical driver for purchasing decisions, brand loyalty, and overall market success. In contrast, misleading, exaggerated, or unsubstantiated claims not only undermine consumer trust but also expose companies to regulatory scrutiny, legal penalties, and reputational harm \cite{dumville2009research}. To safeguard consumers and ensure fair competition, regulatory bodies such as the Federal Trade Commission (FTC) emphasize that product claims, particularly health-related ones, must be truthful, not misleading, and substantiated by robust scientific evidence \cite{FTC2022}. Research consistently highlights that a significant proportion of product advertisements lack accurate or sufficient scientific citations to support their claims \cite{dumville2009research}. 

Product claims need to be legally compliant and scientifically supported. Ideally, manufacturers and marketers also want these claims to be resonant with trendy consumer talk points. Creating these claims requires substantial time and funding as it generally needs to conduct many iterations of designs and tests. Traditionally, the first step is to search a large volume of existing claims and visuals to decide whether one should pivot the existing claims or design brand new claims. These manually crafted candidate textual claims and visuals are then tested with consumers to gauge appeal and collect feedback. After a few iterations, the top appealing claims and visuals are further investigated for scientific support and legal approval. This process can take weeks or even months and require substantial financial resources.   

The gap between compliant and resonant claims and industry practice underscores the urgent need for credible and evidence-based strategies to create claims. Consumers often place disproportionate trust in products labeled as “scientifically studied” or “clinically proven” even when the claims are exaggerated and weakly supported \cite{murphy2023clinically}. This phenomenon reveals not only a vulnerability in consumer decision-making but also a responsibility for marketers and product manufacturers to ensure the integrity and transparency of their product claims.

Recent advances in LLMs offer a promising opportunity to address the above time and resource challenges of creating product claims. LLMs have demonstrated remarkable capabilities in semantic understanding \cite{zhou2024llm} and natural language generation tasks \cite{li2024pre}, and have even exhibited creativity in domains traditionally thought to require human ingenuity \cite{ding2023fluid}. As an example, LLMs have been combined with interactive agents to enable believable simulations of collective human behavior \cite{park2023generative}. 

We developed a platform called \textbf{Claim Advisor} to bridge the gap between LLMs and creation of product claims. \textbf{Claim Advisor} was designed as a minimum-viable-prototype (MVP) web application to quickly search, generate, optimize and rank product claims using LLMs. First, we employ prompt engineering and in-context learning techniques, utilizing previously obtained market research insights to guide the LLM in generating more targeted, persuasive, and compliant claims. This enables a more efficient ideation process while aligning generated claims with real-world consumer expectations and regulatory requirements.

Second, to further streamline the market research phase and reduce its associated costs, we fine-tuned a lightweight version of Microsoft's Phi-3 model \cite{abdin2024phi} using Low-Rank Adaptation (LoRA) techniques \cite{hu2022lora}. This fine-tuned model is designed to simulate initial consumer feedback, allowing for faster iteration cycles before engaging in full-scale market research. Together, these two components create a more agile, cost-effective framework for creating product claims, paving the way for accelerated product development without compromising scientific rigor or consumer trust. 

The code base, dummy data files and example prompts are shared in this  \href{https://github.com/zhy5186612/GenAI-ClaimAdvisor.git}{GitHub repository} to facilitate reapplication and to encourage further research.  

\section{Method}
\subsection{Technical Diagram}
\begin{figure}[htbp]
    \centering
    \includegraphics[width=\columnwidth]{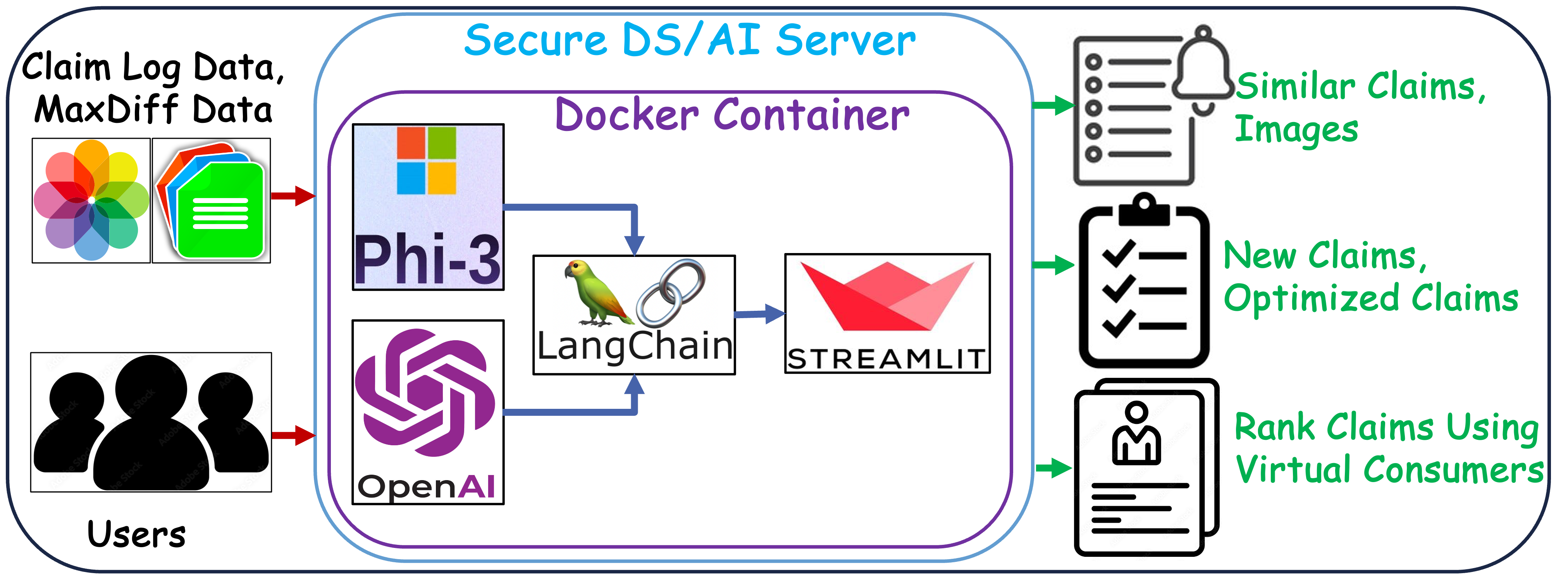}
    \caption{Technical diagram of \textbf{Claim Advisor} MVP web application}
    \label{fig:system_architecture}
\end{figure}

As shown in the figure~\ref{fig:system_architecture}, \textbf{Claim Advisor} takes Claim Log data and Maximum Difference Scaling (MaxDiff) data \cite{marley2005some} as input. Users can input product descriptions, consumer profiles, trending consumer topics and/or visuals to search, generate, optimize and rank claims. Claim Log data contain legally approved claims. It has claim log IDs to link to the related technical support documents. MaxDiff data contain preferences of candidate claims and visuals already screened through MaxDiff studies. MaxDiff is the major research method to study consumers’ preferences on candidate claims before seeking approval.

We used the commercial LLM ChatGPT-4o through Azure for in-context learning and the open source LLM Phi-3 \cite{abdin2024phi} for fine-tuning in our platform. As an MVP web application, \textbf{Claim Advisor} used \textit{LangChain} \cite{langchain} and \textit{streamlit} \cite{streamlit} to build the backend and front end, respectively. Docker \cite{docker} is used to containerize the backend and front end codes and deployed on a secure server. Readers can check the graphical user interface in the appendix \ref{apd: MaxDiff}.   

\subsection{Search Claims}    
To enable semantic claim retrieval from the dataset, we employ text embedding techniques using OpenAI’s TEXT-EMBEDDING-ADA-002 model. Semantic similarity is computed via cosine similarity between the embedded representations of textual inputs. Given the close relationship between claim content and visual design elements, we further support the retrieval of related images by leveraging the CLIP model \cite{radford2021learning}, which aligns text and image embeddings within a shared latent space. This alignment allows user-provided textual inputs to effectively match with relevant visual content and vice versa. To enhance flexibility, we introduce a multimodal fusion approach wherein users can simultaneously input both text and image queries. By specifying a weight parameter $W$, users control the relative influence of textual ($emb_{txt}$) and visual ($emb_{img}$) components. The fused embedding is computed as
\begin{equation}
emb = (1 - W) \cdot emb_{txt} + W \cdot emb_{img}.
\end{equation}
The resulting representation is then used to retrieve the most semantically similar images from existing designs based on cosine similarity in the embedding space. 

If the retrieved claim text and images are from Claim Log data, because claim text and visuals in Claim Log data have been legally approved and have technical support documents, they can be directly repurposed in another usage scenario such as a similar product variant or same product in different distribution channels. If the retrieved contents are from MaxDiff data, they needs to be go through technical tests and legal review before usage on a product.    

In the context of product claim creation, MaxDiff offers a robust methodological framework for empirically assessing consumer preferences among a large number of candidate claims. Given a pool of generated claims \( \{C_1, C_2, \dots, C_N\} \), MaxDiff analysis can identify which claims maximize perceived relevance, credibility, and persuasiveness according to the target audience. This approach overcomes limitations inherent to traditional rating-scale evaluations, where respondents often rate multiple claims similarly, masking meaningful preference differences. Readers can check appendix \ref{apd: MaxDiff} for further description of the MaxDiff method. 

\subsection{Generate and Optimize Claims}\label{sec: GenerateOptimizeClaims}
Prompt engineering and in-context learning are two key techniques for adapting LLMs to specific tasks without requiring fine-tuning. Prompt engineering~\cite{chen2023unleashing} involves carefully designing input prompts to guide the model’s output toward desired behaviors, often by including detailed instructions, task-specific context, or illustrative examples. In-context learning~\cite{dong2022survey} refers to the model’s ability to infer patterns or strategies directly from examples presented within the prompt, allowing it to perform new tasks without updating model parameters.

By leveraging prompt engineering and in-context learning, LLMs can be steered toward producing higher-quality, contextually appropriate outputs, even in specialized domains such as product claim generation. To incorporate knowledge from MaxDiff studies into the LLM, we combined prompt engineering with in-context learning. The prompt includes both target consumer information and examples derived from previous MaxDiff research. The target consumer information is embedded in the system message, while the examples from past MaxDiff studies are incorporated through an in-context learning approach. 

\begin{figure}[htbp]
    \centering
    \includegraphics[width=\columnwidth]{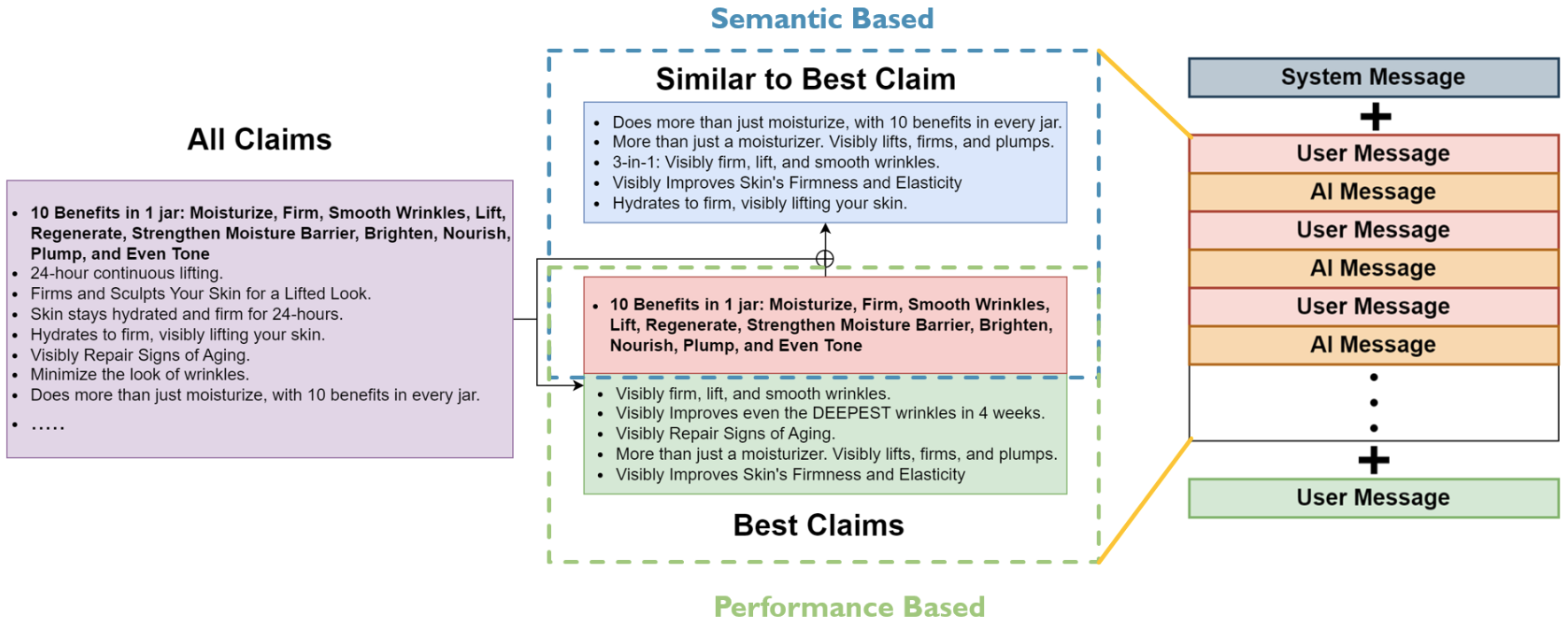}
    \caption{In-context learning examples construction for generating an optimized claim.}
    \label{fig:claim_optimization}
\end{figure}

For each MaxDiff study, we generated examples where the model produces an optimized claim based on five given claims. The examples are constructed using two methods, as illustrated in Figure~\ref{fig:claim_optimization}. The performance based method (indicated by the green dotted line) is based on MaxDiff scores. Our hypothesis is that the LLM can infer consumer preferences by analyzing a set of moderately successful claims and then synthesizing a better, novel claim by combining information from those examples. In our implementation, we selected the second to sixth highest scoring claims as input examples and tasked the model to generate a claim that outperforms them. 

The semantic based method (indicated by the blue dotted line) relies on semantic similarity. The hypothesis for this approach is that claims semantically similar to the best-performing claim may indicate a correct thematic direction. Therefore, we generated examples by selecting the five claims most semantically similar to the best-performing claim. Semantic similarity between claims is calculated using the cosine similarity between embeddings. We generated 300 examples from past MaxDiff studies and used them as in-context learning for claim generation and optimization. The input to both methods are the claims from past MaxDiff studies. The output are the in-context learning examples used in the prompt.    

In practice, both methods can be combined within the same prompt to enhance in-context learning and improve claim generation. To evaluate the model’s performance, we conducted three rounds of MaxDiff research of 30 claims. In the first round, all claims were manually created by human experts. In the second round, new claims were generated by combining previous MaxDiff results with findings from the first round, using the two methods described above. In the third round, claims were generated similarly, based on results from the second round. The number of claims in each round was kept the same (i.e. 30 claims) as in the initial round. Through this iterative process, we aim to progressively optimize the quality of claims by leveraging knowledge accumulated from prior MaxDiff research.

\subsection{Rank Claims}

To virtually screen claims before performing actual MaxDiff research, we guide the model to mimic the MaxDiff process by selecting the best and worst claims from a set of five given claims. We applied prompt engineering and in-context learning techniques, building on the experience from the claim optimization task. We tested model performance under different settings by varying the number of in-context learning examples provided (1, 5, and 10 examples) with a fine-tuned model.

We fine-tuned a lightweight pretrained model, Phi-3 from Microsoft \cite{abdin2024phi}, using the Low-Rank Adaptation (LoRA)~\cite{hu2022lora} method. We evaluated models with 7B and 14B parameters. Fine-tuning refers to the process of updating a pretrained model’s parameters on a small, task-specific dataset to adapt it to new objectives while preserving its general language capabilities. LoRA is a parameter-efficient fine-tuning technique that introduces trainable low-rank matrices into certain layers of the model, significantly reducing computational costs and memory usage. By applying LoRA to Phi-3, we aim to enhance the model’s ability to understand and rank claims according to consumer preferences without requiring full retraining. 

\begin{figure}[htbp]
    \centering
    \includegraphics[width=\columnwidth]{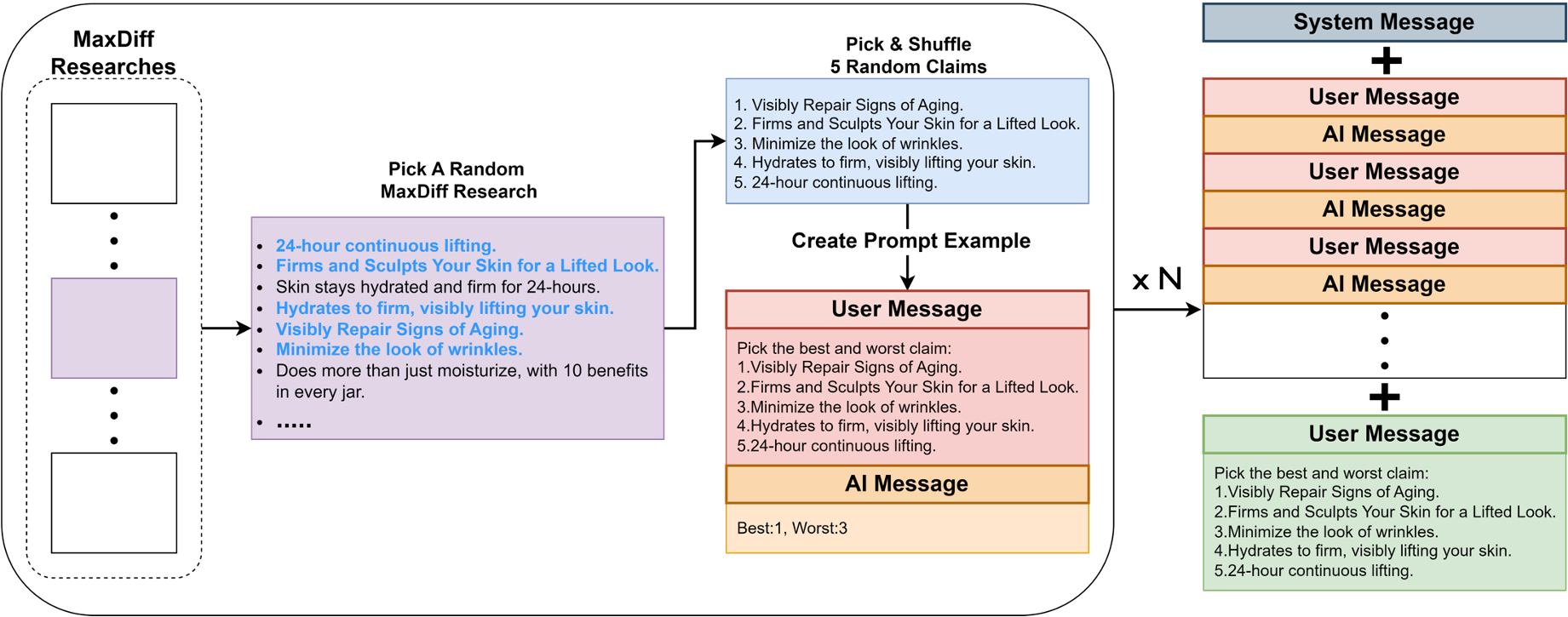}
    \caption{Example construction method for fine-tuning in the claim ranking task.}
    \label{fig:claim_ranking}
\end{figure}

The examples used for fine-tuning are illustrated in Figure~\ref{fig:claim_ranking}. Specifically, we select five claims with known performance and construct the expected output by identifying the best and worst claims. We used 100316 training examples from past MaxDiff studies for fine-tuning. 

During fine-tuning, we also included a single training example as an in-context learning example in the prompt. We observed that including this example helped the model learn the output format more effectively. After obtaining the fine-tuned model, we utilized it by providing five random claims—drawn from the set of claims to be tested—and asking the model to select the best and worst claim. By repeating this process multiple times, we calculated the number of times each claim was selected as the best or worst. We then computed a score for each claim using the ratio of number of times in which it was selected as the best to that of being selected as the worst. These scores were used to rank the generated claims (model-predicted rank). 

We employed Kendall’s tau coefficient~\cite{kendall1938new} to quantitatively evaluate the alignment between model-predicted rank and the true rank. The true rank was obtained through the holdout MaxDiff studies. Kendall’s tau is a non-parametric statistic that measures the ordinal association between two ranked variables by comparing the number of concordant and discordant pairs. Formally, given two rankings, $\tau$ is calculated as:

\begin{equation}
\tau = \frac{C - D}{\frac{1}{2}n(n-1)}
\end{equation}

where $C$ is the number of concordant pairs, $D$ is the number of discordant pairs, and $n$ is the total number of ranked items. A $\tau$ value of $1$ indicates perfect agreement between the rankings, $-1$ indicates complete disagreement, and $0$ suggests no correlation. By using Kendall’s tau, we can robustly assess how well the model’s predicted claim rankings approximate the ground truth without being overly sensitive to minor ranking variations.

\section{Result}
\subsection{Search Claims}
\textbf{Claim Advisor} accelerates searches of text and image of claims, maximizing the return on claim log and MaxDiff study data assets. It can return similar claims within seconds from a large volume of MaxDiff studies and claim log data. Users can use different filters such as age groups and product lines to limit their searches. 

\subsection{Generate and Optimize Claims}
\textbf{Claim Advisor} can generate and/or optimize tens to hundreds of candidate claims within minutes using in-context learning method as detailed in section \ref{sec: GenerateOptimizeClaims}. We evaluate the performance by applying two cutoff percentages to the MaxDiff preference likelihoods, separating 30 claims into three regions: \emph{highly appealing}, \emph{appealing}, and \emph{less appealing}. The preference likelihood of a generated claim is the ratio of times in which it was selected as the best claim to all times in which it was selected as a candidate in a random draw in a round of a MaxDiff study. The two cutoff percentages $P1$ and $P2$ ($P2$ > $P1$) are determined empirically based on our internal MaxDiff studies. The results are shown in Table~\ref{tab:claim_opt_res}. If the preference likelihood is great than $P2$, it is classified as highly appealing. If it is less than or equal to $P1$, it is less appealing.      

In the first round (claims designed by human experts), the majority of claims (46\%) fall into the appealing region, while 20\% of claims are highly appealing to consumers and 33\% are categorized as less appealing. In the second round, we observed an improvement: the proportion of highly appealing claims increases by 13\%, reaching 33\%, while the proportion of less appealing claims decreases from 34\% to 31\%. 

By the third round, all claims achieve highly appealing performance. This surprisingly strong result demonstrates that the model, even relying only on in-context learning and prompt engineering, can effectively learn consumer preferences within a relatively small number of iterations. These findings highlight the strong potential of LLMs to assist humans in creative tasks such as claim creation. 

\begin{table}
  \caption{Performance of generated claims in three rounds of MaxDiff researches. }
  \label{tab:claim_opt_res}
  \centering
  \begin{tabular}{lccc}
    \toprule
    & Round-1 (Human)     & Round-2 (Claim Advisor) & Round-3 (Claim Advisor) \\
    \midrule
    High Appealing & 20\% & 33\% & 100\% \\
    Appealing      & 46\% & 36\% &   0\% \\
    Less Appealing & 34\% & 31\% &   0\% \\
    \bottomrule
  \end{tabular}
\end{table}

\subsection{Rank Claims}
The performance of the claim ranking models was evaluated using Kendall’s tau coefficient. We compared our fine-tuned models against ChatGPT-3.5, GPT-4, and GPT-4o\footnote{Accessed via Azure OpenAI API in July 2024}. To ensure a fair comparison between our fine-tuned models and the non-fine-tuned ChatGPT models, we provided the ChatGPT models with 100 examples from previous MaxDiff studies, whereas our models received only a small number of examples (1, 5, or 10). This setup helps minimize differences in the amount of contextual information provided to each model.

\begin{figure}[htbp]
    \centering
    \includegraphics[width=0.8\columnwidth]{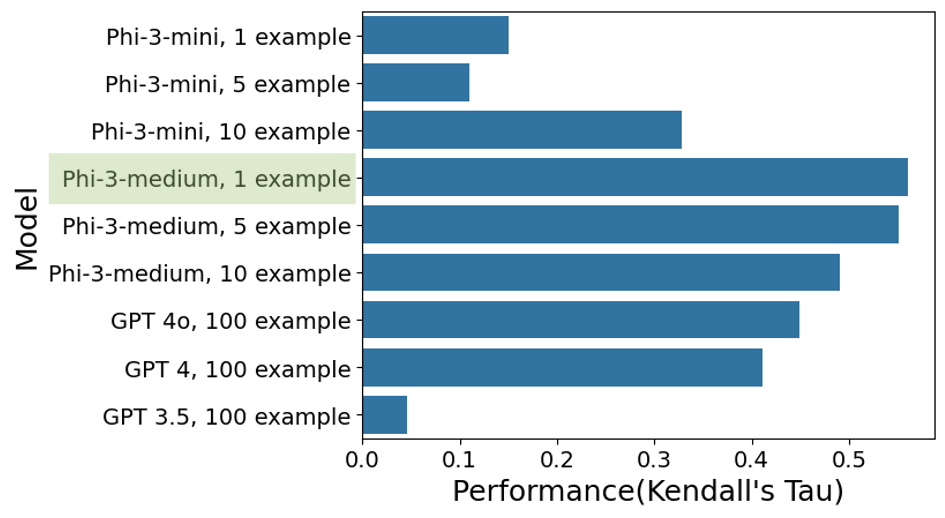}
    \caption{Model performance based on Kendall’s tau. \textit{Phi-3-mini} and \textit{Phi-3-medium} denote models with 7B and 14B parameters, respectively. \textit{GPT} refers to ChatGPT models from OpenAI. Best performed model is marked with green box.}
    \label{fig:claim_ranking_kendalls_tau}
\end{figure}

Figure~\ref{fig:claim_ranking_kendalls_tau} presents Kendall’s tau coefficients across all evaluated models. The Phi-3 model with 14B parameters consistently outperforms all others. Although there is a noticeable improvement in performance from GPT-3.5 to GPT-4 and GPT-4o, none of the ChatGPT models surpass our best-performing model. Notably, the Phi-3 7B model (denoted as \textit{mini}), despite its smaller size, achieves performance comparable to the 14B version (\textit{medium}) when using only 10 in-context examples. It also significantly outperforms GPT-3.5, even though the latter was given 10 times more examples.

Interestingly, increasing the number of in-context examples does not always improve performance. For instance, the Phi-3 medium model shows a slight performance drop when moving from one to ten examples. Overall, the best result was achieved by the Phi-3 medium model using only a single example, suggesting that smaller, well-designed prompts can be highly effective for this task.

\begin{figure}[htbp]
    \centering
    \includegraphics[width=0.8\columnwidth]{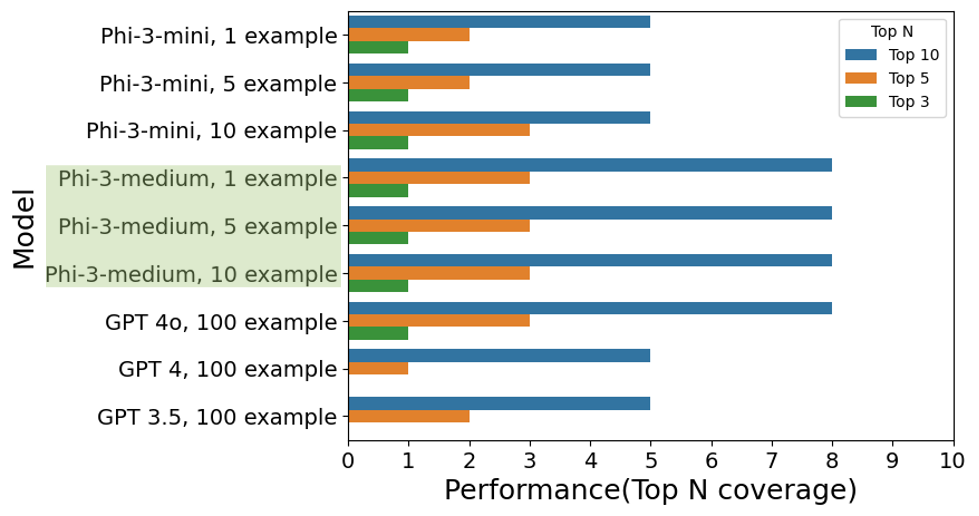}
    \caption{Top-N coverage performance. \textit{Phi-3-mini} and \textit{Phi-3-medium} denote models with 7B and 14B parameters, respectively. \textit{GPT} refers to ChatGPT models from OpenAI. Best performed model is marked with green box.}
    \label{fig:claim_ranking_top_n}
\end{figure}

To further analyze model utility in practical applications, we measured top-N coverage, shown in Figure~\ref{fig:claim_ranking_top_n}. This metric evaluates how many of the model’s top-N predicted claims align with the top-N claims identified through actual MaxDiff research. This is particularly relevant when users are only interested in identifying the most promising claims, rather than perfectly ordering all candidates.

We observed that the Phi-3 mini model, despite its lower Kendall’s tau score, achieves top-3 and top-5 coverage comparable to the Phi-3 medium model when provided with 10 examples. In contrast, GPT-3.5 and GPT-4 exhibit the lowest top-N coverage, with both failing to capture any of the top 3 claims. GPT-4o performs better, matching our best model when provided with 100 in-context examples. Nevertheless, our fine-tuned Phi-3 models—especially given their smaller size and fewer required examples—demonstrate superior efficiency and scalability for claim ranking tasks, offering substantial cost and resource advantages.

\section{Discussion}

For claim optimization, we found that prompt engineering contributed to noticeable performance improvements. Specifically, incorporating domain knowledge into prompts helped guide the LLMs to generate claims more closely aligned with interests of targeted consumers. However, we also observed a trade-off: excessive instruction in prompts can reduce output diversity, leading to repetitive claims centered on the provided guidance. These findings highlight that while prompt design remains an underexplored area, well-crafted prompts can significantly influence the effectiveness of LLM-driven generation tasks. Diverse in-context learning examples from multiple product categories may help to mitigate the issue of lack of diversity caused by excessive instruction. Further experiments need to be conducted to investigate the robustness of claim generation and optimization based on the number of input claims, the cut off percentages and the number of in-context learning examples.    

For claim ranking, our initial approach involved asking the LLM to rank all claims in a single pass. However, we found that the resulting rankings lacked statistical alignment with MaxDiff outcomes. By mimicking the MaxDiff methodology, namely asking the model to select the best and worst claims from small sets, we achieved results that were statistically meaningful and closely matched real-world consumer preferences. This underscores a current limitation of LLMs: while capable of generating plausible outputs, they may struggle with tasks that require implicit statistical reasoning unless they are structured in a way that reflects the underlying process. These bias in LLM have been well documented and could be mitigated by technics such as contextual calibration procedure \cite{zhao2021calibrate}.  

In a MaxDiff study, a consumer is asked to pick the best and the worst in a set randomly picked from a pool. It does not need to provide ranks for every items in the set. It is a nice balances between ranking every items in a set (hard to differentiate all items) and only picking the best item (losing what a consumer does not like). MaxDiff plays into psychology and recognition when choosing a product from a long list of offerings. In practice, we found MaxDiff has been very effective to screen concepts, claims etc.  

Our experiments demonstrate that effective system design, which integrates traditional methods and domain expertise, can enable LLMs to perform complex tasks with high reliability. Using open-source models such as Phi-3 offers additional benefits, including greater control, transparency, and lower inference costs. In particular, our fine-tuned lightweight models outperformed larger commercial alternatives such as GPT-3.5 and GPT-4, despite using significantly fewer in-context examples. We did not compare a model with only prompt (no in-context examples) with a fine-tuned model because early tests showed LLM prompt alone tends to generate generic claims without specificity. 

Several practical considerations emerged from our deployment experience. First, latency remains a challenge, particularly for claim ranking, which requires thousands of inferences within a short period of time. Second, the stability of the model is critical. OpenAI's commercial models, while powerful, are periodically updated, which can affect reproducibility and prompt behavior. In contrast, open-source models offer greater consistency and can be fine-tuned to maintain performance over time. Finally, cost management is more feasible with open source models, especially when lightweight fine-tuned versions can match or exceed the performance of commercial offerings.

Overall, our work demonstrates the strong potential of LLMs in assisting human creativity and decision-making tasks, provided that their use is carefully designed and informed by domain knowledge. We believe our pipeline could generalize on usage cases from other industries given product claims are needed virtually everywhere for either physical or digital products. We share the code base and hope to encourage research, reapplication and improve the pipeline from academia and industry.

\section{Conclusion}
A MVP web application \textbf{Claim Advisor} has been developed to accelerate product claim creations using in-context learning and fine-tuning of LLMs. It leverages LLMs to learn from existing claim log data, MaxDiff research data and domain expertise to disrupt the speed and economics of claim search, generation, optimization, and simulation. The application in a consumer packaged goods company has significantly accelerated the efficiency and creativity of product researchers in creating product claims.   

\section{Limitations}

This study has several limitations that impact the transparency and reproducibility of our findings. First, our research relies on a proprietary dataset derived from internal MaxDiff studies, which cannot be publicly released due to confidentiality constraints. Additionally, the specific prompts used in our experiments are also not fully disclosed, as they contain proprietary information closely tied to the dataset and internal business logic. Furthermore, while we fine-tuned a lightweight open-source model (Phi-3), the resulting model cannot be publicly shared due to the use of non-disclosable training data. 

For the evaluation of commercial LLMs such as ChatGPT-3.5, GPT-4, and GPT-4o, results were obtained via the OpenAI API through Azure. Because these models are closed-source and subject to updates without notice, the outcomes are not guaranteed to be reproducible. Lastly, the interactive Web-GUI system developed as part of this research (\textbf{Claim Advisor}) is currently not published or available for public use. 

Together, the limitations above restrict the replicability and open access of our research. However, we shared our code base, example prompt and necessary dummy data files to show the format of input data in this \href{https://github.com/zhy5186612/GenAI-ClaimAdvisor.git}{GitHub repository}. Researchers are encouraged to reapply using their own data and to conduct further research in this area based on our code base.

\begin{ack}
We thank our colleagues Kelly Anderson, Gabriel Comeron, Weronika Koga, George Gabone, Matt Barker and Oya Aran for their help and valuable feedback during the research and deployment of \textbf{Claim Advisor}. This research is solely sponsored by P\&G. Authors have no conflict of interest.
\end{ack}

\bibliographystyle{abbrv}
\bibliography{ref}




\newpage
\appendix
\section{Web-Based GUI} \label{apd: GUI}
\begin{figure}[htbp]
    \centering
    \includegraphics[width=0.85\columnwidth]{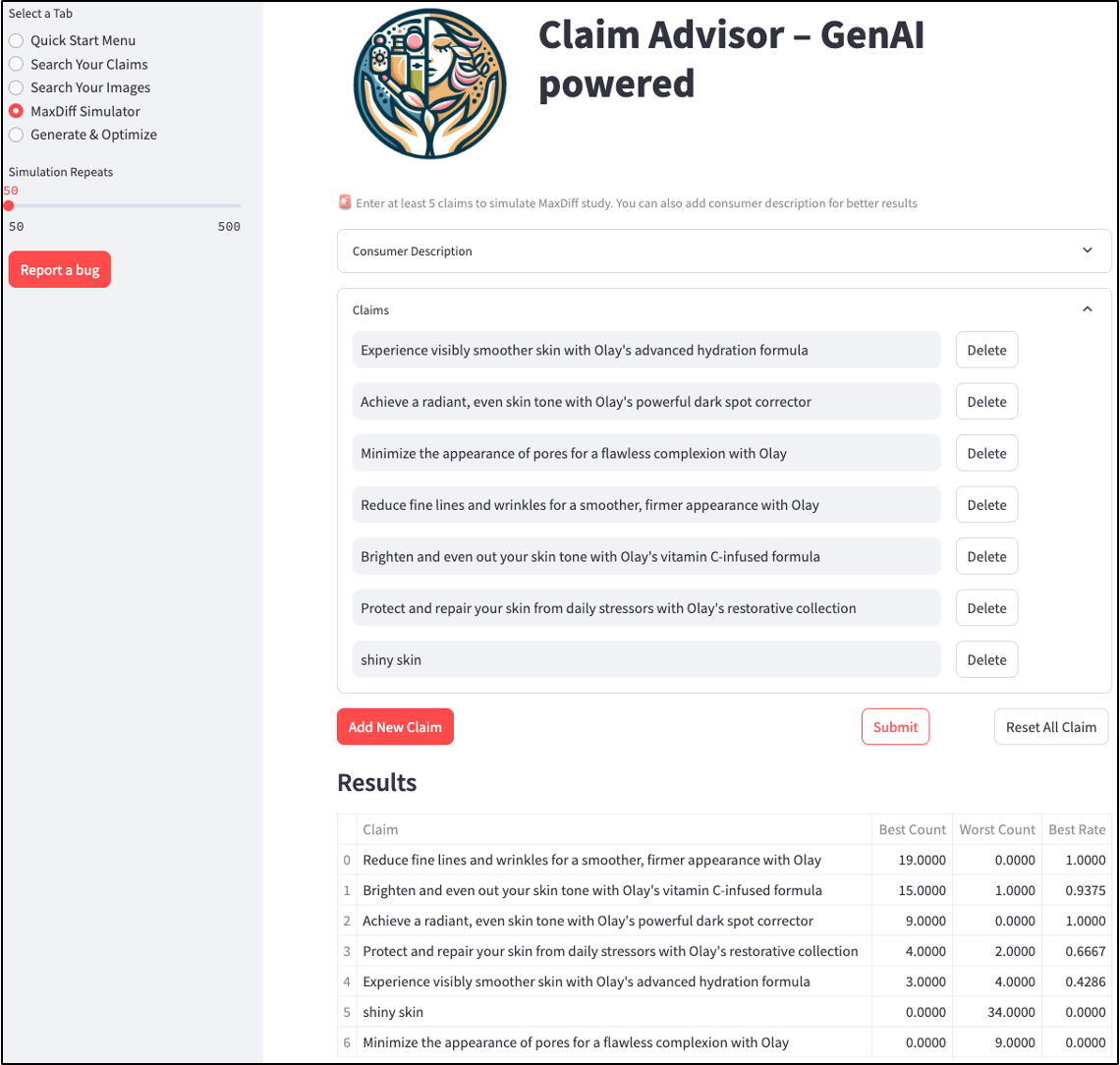}
    \caption{Graphical user interface of the \textbf{Claim Advisor} MVP web application.}
    \label{fig:caGUI}
\end{figure}

To bridge the gap between LLMs and end users, we developed a web-based graphical user interface (GUI) system, referred to as \textbf{Claim Advisor}. This MVP web application was deployed on a GPU server on premise ( NVIDIA GeForce RTX 2080, 11GB memory).
It allows users to search existing claims, generate and optimize new claims and perform preliminary screening using a simulated MaxDiff study. Figure~\ref{fig:caGUI} shows a snapshot of the GUI. It shows that "MaxDiff Simulator" is used to rank 6 generated claims and 1 manually created claim (shiny skin) using 50 round of simulations.

\textbf{Claim Advisor }is implemented as a secure containerized pipeline for the automated search, generation, optimization and evaluation of product claims (Figure~\ref{fig:system_architecture}). The system is deployed within a \textit{Docker} container\cite{docker} on a secure server. It ingests structured inputs, such as historical claim logs and MaxDiff data, which reflect consumer preference patterns. Users interact with the system via a MVP web application built using \textit{Streamlit}\cite{streamlit}. Internally, the workflow is orchestrated by \textit{LangChain}\cite{langchain}, which coordinates the flow of data and manages interactions across multiple AI components. 

The MVP web application of \textbf{Claim Advisor } has been productionized at a consumer packaged goods company. It has significantly accelerated product researchers efficiency and creativity to search, generate, optimize and rank product claims.    

\section{Maximum Difference Scaling} \label{apd: MaxDiff}
Maximum Difference Scaling (MaxDiff)\cite{marley2005some}, also known as best-worst scaling, is a discrete choice modeling technique used to measure relative preferences or importance scores among a set of items. In a typical MaxDiff task, given a subset \( S \subseteq \{i_1, i_2, \dots, i_k\} \) of \( k \) items selected from a larger item pool of size \( N \), respondents are asked to choose the ``best'' (most preferred) and ``worst'' (least preferred) items. \( k \) and \( N \) can typically take values of 3-6 and 20-100, respectively. Each choice provides information about the underlying utility difference between items: if item \( i \) is selected as best and item \( j \) as worst, it implies $U(i) - U(j) > U(m) - U(n)$ for all other pairs \( (m, n) \) in \( S \), where \( U(\cdot) \) represents the latent utility associated with each item. By aggregating responses across multiple sets and respondents, it is possible to estimate the utilities \( U(i) \) for all items using models such as multinomial logit (MNL) or hierarchical Bayesian estimation. These utilities can then be normalized (e.g., to sum to 100) to produce a ratio-scaled measure of relative importance\cite{marley2005some}.

\end{document}